\documentclass[10pt,a4paper,openbib]{article}
\usepackage[latin1]{inputenc}
\usepackage{amsmath, amsthm}
\usepackage{amsfonts}
\usepackage{amssymb}
\usepackage{graphicx}
\usepackage[colorlinks=false,
            linkcolor=blue,
            anchorcolor=blue,
            citecolor=green,
            ]{hyperref}
\usepackage{xcolor}
\usepackage{algorithm}
\usepackage{algorithmic}
\usepackage{cite}
\usepackage{indentfirst}
\usepackage{etoolbox}
\begin{document}
\title{Equivalence between LINE and Matrix Factorization}
\author{
Qiao Wang, Zheng Wang\thanks{Corresponding author: Zheng Wang.}, Xiaojun Ye\\
School of Software, Tsinghua University, Beijing, China\\
\{wangqiao15, zheng-wang13\}@mails.tsinghua.edu.cn \\
\{yexj\}@tsinghua.edu.cn
}
\maketitle

\begin{abstract}
LINE~\cite{tang2015line}, as an efficient network embedding method, has shown its effectiveness in dealing with large-scale undirected, directed, and/or weighted networks.
Particularly, it proposes to preserve both the local structure (represented by First-order Proximity) and global structure (represented by Second-order Proximity) of the network.
In this study, we prove that LINE with these two proximities (LINE(1st) and LINE(2nd)) are actually factoring two different matrices separately.
Specifically, LINE(1st) is factoring a matrix $M^{(1)}$, whose entries are the doubled Pointwise Mutual Information (PMI) of vertex pairs in undirected networks, shifted by a constant.
LINE(2nd) is factoring a matrix $M^{(2)}$, whose entries are the PMI of vertex and context pairs in directed networks, shifted by a constant.
We hope this finding would provide a basis for further extensions and generalizations of LINE.
\end{abstract}

\section{Notation and Definition}
Given a network $\mathcal G=(\mathcal V,\mathcal E)$, where each edge $e\in \mathcal E$ is an ordered pair $e=(v_{i},v_{j})$ and has an associated weight $w_{ij}>0$.
In directed networks, the in-degree and out-degree of vertex $v_{i}$ are denoted as $deg^{-}(v_{i})$ and $deg^{+}(v_{i})$ respectively.
In addition, in undirected networks, the degree of vertex $v_{i}$ is denoted as $deg(v_{i})$.

The first-order proximity~\cite{tang2015line} characterizes the local structure similarity between vertices. More specifically, if $(v_{i},v_{j})\in \mathcal E$, $w_{ij}$ indicates the first-order proximity between $v_{i}$ and $v_{j}$, otherwise their first-order proximity is 0.

The second-order proximity~\cite{tang2015line} characterizes the global structure similarity between vertices.
Mathematically, let $p_{i} = (w_{i1},...,w_{i|\mathcal V|})$ represent the first-order proximity between $v_{i}$ and the other vertices, then the second-order proximity between $v_{i}$ and $v_{j}$ is characterized by the similarity between $p_{i}$ and $p_{j}$.

To simultaneously preserve these two proximities, Tang et al.~\cite{tang2015line} train the LINE model which preserves the first-order proximity (denoted as LINE(1st)) and second-order proximity (denoted as LINE(2nd)) separately and then concatenate the embeddings learned by these two methods.

\section{Proof}
Levy and Goldberg~\cite{Omer2014IMF} have shown that Skip-Gram with Negative Sampling~\cite{Mikolov2013DRW} is implicitly factoring a word-context matrix.
Similarly, in this study we prove that LINE(1st) and LINE(2nd) are actually factoring two different matrices separately.
For ease of understanding, we first give the proof of LINE(2nd) and then give the proof of LINE(1st).

\subsection{Equivalence of LINE(2nd) and Matrix Factorization}
LINE(2nd) assumes the given network is directed (an undirected edge can be treated as two directed edges with opposite directions and equal weights), and for each directed edge $(v_{i},v_{j})$ it defines $v_{j}$ as the ``context'' of $v_{i}$.
As such, on one hand each vertex $v_{i}\in \mathcal V$ is embedded into a d-dimensional vector $\overrightarrow{v_{i}}$ ($d\ll |\mathcal V|$) when it plays as the vertex itself;
on the other hand it is embedded into a d-dimensional vector $\overrightarrow{u_{i}}$ when it plays as a ``context'' of the others.
Let $V$ denote a $d\times |\mathcal V|$ matrix whose $i$-th column is the vertex embedding $\overrightarrow{v_{i}}$ and
$U$ denote a $d\times |\mathcal V|$ matrix whose $j$-th column is the ``context'' embedding $\overrightarrow{u_{j}}$.
We will figure out that LINE(2nd) is factoring a matrix $M^{(2)}=V^{T}U$.

According to~\cite{tang2015line}, LINE(2nd) minimizes the following objective function:
\begin{equation}\label{second_order_objective}
\begin{aligned}
O_{2}=-\sum_{(i,j)\in \mathcal E} w_{ij}\log p_{2}(v_{j}|v_{i})=-\sum_{(i,j)\in \mathcal E} w_{ij}\log \frac{exp({\overrightarrow{u_{j}}}^{T}\cdot \overrightarrow{v_{i}})}{\sum\nolimits_{k=1}^{|\mathcal V|}exp({\overrightarrow{u_{k}}}^{T}\cdot \overrightarrow{v_{i}})}
\end{aligned}
\end{equation}

Optimizing Eq.~\ref{second_order_objective} is time-consuming, since it requires the summation over all vertices when calculating the conditional probability $p_{2}(\cdot|v_{i})$.
Therefore, LINE(2nd) adopts the negative sampling approach~\cite{Mikolov2013DRW}, which replaces each $\log p_{2}(v_{j}|v_{i})$ in Eq.~\ref{second_order_objective} with the following objective function:
\begin{equation}\label{second_order_ns}
\begin{aligned}
\log \sigma({\overrightarrow{u_{j}}}^{T}\cdot \overrightarrow{v_{i}}) + k\cdot E_{v_{n}\sim P_{n}(v)}\log \sigma(-{\overrightarrow{u_{n}}}^{T}\cdot \overrightarrow{v_{i}}),
\end{aligned}
\end{equation}
where $\sigma(x)=1/(1+exp(-x))$ is the sigmoid function, $P_{n}(v)$ is a negative sampling distribution, and $k$ defines the number of negative edges.

Next, by substituting Eq.~\ref{second_order_ns} into Eq.~\ref{second_order_objective}, we can rewrite the objective function of LINE(2nd) as:
\begin{equation}\label{second_order_rewrite}
\begin{aligned}
O_{2}&=-\sum_{(i,j)\in \mathcal E} w_{ij}[\log \sigma({\overrightarrow{u_{j}}}^{T}\cdot \overrightarrow{v_{i}}) + k\cdot E_{v_{n}\sim P_{n}(v)}\log \sigma(-{\overrightarrow{u_{n}}}^{T}\cdot \overrightarrow{v_{i}})] \\
&=-\sum_{(i,j)\in \mathcal E} w_{ij}\log \sigma({\overrightarrow{u_{j}}}^{T}\cdot \overrightarrow{v_{i}})-\sum_{(i,j)\in \mathcal E} w_{ij}\cdot k\cdot E_{v_{n}\sim P_{n}(v)}\log \sigma(-{\overrightarrow{u_{n}}}^{T}\cdot \overrightarrow{v_{i}}) \\
&=-\sum_{(i,j)\in \mathcal E} w_{ij}\log \sigma({\overrightarrow{u_{j}}}^{T}\cdot \overrightarrow{v_{i}})-\sum_{v_{i}\in \mathcal V}deg^{+}(v_{i})\cdot k\cdot E_{v_{n}\sim P_{n}(v)}\log \sigma(-{\overrightarrow{u_{n}}}^{T}\cdot \overrightarrow{v_{i}})
\end{aligned}
\end{equation}

\newpage
To simplify the analysis, here we set the negative sampling distribution $P_{n}(v)\propto deg^{-}(v_{n})$
\footnote{LINE(2nd) sets the negative sampling distribution $P_{n}(v)\propto {deg^{+}(v_{n})}^{3/4}$. In our proof, we replace it with $P_{n}(v)\propto deg^{-}(v_{n})$.
On one hand, for simplicity, we use the unigram distribution instead of its $3/4$ power following~\cite{Omer2014IMF}.
On the other hand, according to the definition of negative sampling~\cite{Mikolov2013DRW}, the sampled negative edges for $(v_{i},v_{j})$ should have the same starting point (i.e., $v_{i}$).
Therefore, in this proof, we draw negative samples according to the in-degree of vertices.}.
Hence, the expectation term in Eq.~\ref{second_order_rewrite} can be specified as follows:
\begin{equation}\label{second_order_expectation}
\begin{aligned}
&E_{v_{n}\sim P_{n}(v)}\log \sigma(-{\overrightarrow{u_{n}}}^{T}\cdot \overrightarrow{v_{i}})=\sum_{v_{n}\in V} \frac{deg^{-}(v_{n})}{\sum\nolimits_{v_{n}\in \mathcal V} deg^{-}(v_{n})}\cdot \log \sigma(-{\overrightarrow{u_{n}}}^{T}\cdot \overrightarrow{v_{i}}) \\
&=\frac{deg^{-}(v_{j})}{\sum\nolimits_{v_{n}\in \mathcal V}deg^{-}(v_{n})}\cdot \log \sigma(-{\overrightarrow{u_{j}}}^{T}\cdot \overrightarrow{v_{i}})+\sum_{v_{n}\in \mathcal V\backslash \left\{ v_{j} \right\}} \frac{deg^{-}(v_{n})}{\sum\nolimits_{v_{n}\in \mathcal V}deg^{-}(v_{n})}\cdot \log \sigma(-{\overrightarrow{u_{n}}}^{T}\cdot \overrightarrow{v_{i}})
\end{aligned}
\end{equation}

As each product ${\overrightarrow{u_{j}}}^{T}\cdot \overrightarrow{v_{i}}$ is independent with the others, we can gain the local objective for a specific $(v_{i},v_{j})$ pair by combining Eqs.~\ref{second_order_rewrite} and~\ref{second_order_expectation}:
\begin{equation}\label{second_order_local_objective}
\begin{aligned}
\ell(v_{i},v_{j})=w_{ij}\cdot \log \sigma({\overrightarrow{u_{j}}}^{T}\cdot \overrightarrow{v_{i}})+deg^{+}(v_{i})\cdot k\cdot \frac{deg^{-}(v_{j})}{\sum\nolimits_{v_{n}\in \mathcal V}deg^{-}(v_{n})}\cdot \log \sigma(-{\overrightarrow{u_{j}}}^{T}\cdot \overrightarrow{v_{i}})
\end{aligned}
\end{equation}

To minimize the objective function of LINE(2nd) (i.e., Eq.~\ref{second_order_objective}), we must maximize $\ell(v_{i},v_{j})$. As such, we define $x={\overrightarrow{u_{j}}}^{T}\cdot \overrightarrow{v_{i}}$ and get the derivative of $\ell(v_{i},v_{j})$ with respect to $x$:
\begin{equation}\label{second_order_partial}
\begin{aligned}
\frac{\partial \ell}{\partial x}=w_{ij}\cdot \sigma(-x)-deg^{+}(v_{i})\cdot k\cdot \frac{deg^{-}(v_{j})}{\sum\nolimits_{v_{n}\in \mathcal V}deg^{-}(v_{n})}\cdot \sigma(x)
\end{aligned}
\end{equation}

Comparing the derivative to zero, we have:
\begin{equation}\label{second_order_result}
\begin{aligned}
x={\overrightarrow{u_{j}}}^{T}\cdot \overrightarrow{v_{i}}=\log \frac{w_{ij}\cdot \sum\nolimits_{v_{n}\in \mathcal V}deg^{-}(v_{n})}{deg^{+}(v_{i})\cdot deg^{-}(v_{j})}-\log k
\end{aligned}
\end{equation}

Notably, the expression $\log \frac{w_{ij}\cdot \sum\nolimits_{v_{n}\in \mathcal V}deg^{-}(v_{n})}{deg^{+}(v_{i})\cdot deg^{-}(v_{j})}$ is the Pointwise Mutual Information (PMI)~\cite{church1990word} of vertex pair $(v_{i},v_{j})$ in directed networks.

Overall, therefore, we can characterize the matrix $M^{(2)}$ that LINE(2nd) is actually factoring:
\begin{equation}\label{second_order_matrix_factorization}
\begin{aligned}
M^{(2)}_{ij}={\overrightarrow{u_{j}}}^{T}\cdot \overrightarrow{v_{i}}=PMI(v_{i},v_{j})-\log k
\end{aligned}
\end{equation}

\subsection{Equivalence of LINE(1st) and Matrix Factorization}
LINE(1st) is only applicable for undirected networks. Each vertex $v_{i}\in \mathcal V$ is embedded into a d-dimensional vector $\overrightarrow{v_{i}}$ ($d\ll |\mathcal V|$) in this method.
Let $V$ denote a $d\times |\mathcal V|$ matrix whose $i$-th column is $\overrightarrow{v_{i}}$.
We will figure out that LINE(1st) is factoring a matrix $M^{(1)}=V^{T}V$.

According to~\cite{tang2015line}, LINE(1st) minimizes the following objective function:
\begin{equation}\label{first_order_objective}
\begin{aligned}
O_{1}=-\sum_{(i,j)\in \mathcal E} w_{ij}\log p_{1}(v_{i},v_{j})=-\sum_{(i,j)\in \mathcal E} w_{ij}\log \frac{1}{1+exp(-{\overrightarrow{v_{i}}}^{T}\cdot \overrightarrow{v_{j}})}
\end{aligned}
\end{equation}

To avoid the trivial solution, LINE(1st) also uses the negative sampling approach (specified in Eq.~\ref{second_order_ns}) by just replacing $\overrightarrow{u_{j}}$ with $\overrightarrow{v_{j}}$.
More specifically, LINE(1st) replaces each $\log p_{1}(v_{i},v_{j})$ in Eq.~\ref{first_order_objective} with the following objective function:
\begin{equation}\label{first_order_ns}
\begin{aligned}
\log \sigma({\overrightarrow{v_{j}}}^{T}\cdot \overrightarrow{v_{i}}) + k\cdot E_{v_{n}\sim P_{n}(v)}\log \sigma(-{\overrightarrow{v_{n}}}^{T}\cdot \overrightarrow{v_{i}}),
\end{aligned}
\end{equation}
where $\sigma(x)=1/(1+exp(-x))$ is the sigmoid function, $P_{n}(v)$ is a negative sampling distribution, and $k$ defines the number of negative edges.

Next, by substituting Eq.~\ref{first_order_ns} into Eq.~\ref{first_order_objective}, we can rewrite the objective function of LINE(1st) as:
\begin{equation}\label{first_order_rewrite}
\begin{aligned}
O_{1}&=-\sum_{(i,j)\in \mathcal E} w_{ij}[\log \sigma({\overrightarrow{v_{j}}}^{T}\cdot \overrightarrow{v_{i}}) + k\cdot E_{v_{n}\sim P_{n}(v)}\log \sigma(-{\overrightarrow{v_{n}}}^{T}\cdot \overrightarrow{v_{i}})] \\
&=-\sum_{(i,j)\in \mathcal E} w_{ij}\log \sigma({\overrightarrow{v_{j}}}^{T}\cdot \overrightarrow{v_{i}})-\sum_{(i,j)\in \mathcal E} w_{ij}\cdot k\cdot E_{v_{n}\sim P_{n}(v)}\log \sigma(-{\overrightarrow{v_{n}}}^{T}\cdot \overrightarrow{v_{i}}) \\
&=-\sum_{(i,j)\in \mathcal E} w_{ij}\log \sigma({\overrightarrow{v_{j}}}^{T}\cdot \overrightarrow{v_{i}})-\sum_{v_{i}\in \mathcal V}deg(v_{i})\cdot k\cdot E_{v_{n}\sim P_{n}(v)}\log \sigma(-{\overrightarrow{v_{n}}}^{T}\cdot \overrightarrow{v_{i}})
\end{aligned}
\end{equation}

To simplify the analysis, here we set the negative sampling distribution $P_{n}(v)\propto deg(v_{n})$
\footnote{LINE(1st) sets the negative sampling distribution $P_{n}(v)\propto {deg(v_{n})}^{3/4}$. In our proof, for simplicity, we replace it with $P_{n}(v)\propto deg(v_{n})$ following~\cite{Omer2014IMF}.}.
Hence, the expectation term in Eq.~\ref{first_order_rewrite} can be specified as follows:
\begin{equation}\label{first_order_expectation}
\begin{aligned}
&E_{v_{n}\sim P_{n}(v)}\log \sigma(-{\overrightarrow{v_{n}}}^{T}\cdot \overrightarrow{v_{i}})=\sum_{v_{n}\in \mathcal V} \frac{deg(v_{n})}{\sum\nolimits_{v_{n}\in \mathcal V} deg(v_{n})}\cdot \log \sigma(-{\overrightarrow{v_{n}}}^{T}\cdot \overrightarrow{v_{i}}) \\
&=\frac{deg(v_{j})}{\sum\nolimits_{v_{n}\in \mathcal V}deg(v_{n})}\cdot \log \sigma(-{\overrightarrow{v_{j}}}^{T}\cdot \overrightarrow{v_{i}})+\sum_{v_{n}\in \mathcal V\backslash \left\{ v_{j} \right\}} \frac{deg(v_{n})}{\sum\nolimits_{v_{n}\in \mathcal V}deg(v_{n})}\cdot \log \sigma(-{\overrightarrow{v_{n}}}^{T}\cdot \overrightarrow{v_{i}})
\end{aligned}
\end{equation}

As each product ${\overrightarrow{v_{j}}}^{T}\cdot \overrightarrow{v_{i}}$ is independent with others, we can gain the local objective for a specific $(v_{i},v_{j})$ pair by combining Eqs.~\ref{first_order_rewrite} and~\ref{first_order_expectation}:
\begin{equation}\label{first_order_local_objective}
\begin{aligned}
\ell(v_{i},v_{j})=w_{ij}\cdot \log \sigma({\overrightarrow{v_{j}}}^{T}\cdot \overrightarrow{v_{i}})+deg(v_{i})\cdot k\cdot \frac{deg(v_{j})}{\sum\nolimits_{v_{n}\in \mathcal V}deg(v_{n})}\cdot \log \sigma(-{\overrightarrow{v_{j}}}^{T}\cdot \overrightarrow{v_{i}})
\end{aligned}
\end{equation}

To minimize the objective function of LINE(1st) (i.e., Eq.~\ref{first_order_objective}), we must maximize $\ell(v_{i},v_{j})$.
As such, we define $x={\overrightarrow{v_{j}}}^{T}\cdot \overrightarrow{v_{i}}$ and get the derivative of $\ell(v_{i},v_{j})$ with respect to $x$:
\begin{equation}\label{first_order_partial}
\begin{aligned}
\frac{\partial \ell}{\partial x}=w_{ij}\cdot \sigma(-x)-deg(v_{i})\cdot k\cdot \frac{deg(v_{j})}{\sum\nolimits_{v_{n}\in \mathcal V}deg(v_{n})}\cdot \sigma(x)
\end{aligned}
\end{equation}

Comparing the derivative to zero, we have:
\begin{equation}\label{first_order_result}
\begin{aligned}
x={\overrightarrow{v_{j}}}^{T}\cdot \overrightarrow{v_{i}}=\log \frac{w_{ij}\cdot \sum\nolimits_{v_{n}\in \mathcal V}deg(v_{n})}{deg(v_{i})\cdot deg(v_{j})}-\log k
\end{aligned}
\end{equation}

Notably, in undirected networks, the PMI of $(v_{i},v_{j})$ is $\log \frac{w_{ij}\cdot \sum\nolimits_{v_{n}\in \mathcal V}deg(v_{n})}{2\cdot deg(v_{i})\cdot deg(v_{j})}$.
Consequently, there exists the following relationship:
\begin{equation}\label{first_order_doubled_PMI}
\begin{aligned}
\log \frac{w_{ij}\cdot \sum\nolimits_{v_{n}\in \mathcal V}deg(v_{n})}{deg(v_{i})\cdot deg(v_{j})}=2PMI(v_{i},v_{j})
\end{aligned}
\end{equation}

Overall, therefore, we can characterize the matrix $M^{(1)}$ that LINE(1st) is actually factoring:
\begin{equation}\label{First_order_matrix_factorization}
\begin{aligned}
M^{(1)}_{ij}={\overrightarrow{v_{j}}}^{T}\cdot \overrightarrow{v_{i}}=2PMI(v_{i},v_{j})-\log k
\end{aligned}
\end{equation}

\bibliographystyle{unsrt}
\bibliography{Equivalence_between_LINE_and_Matrix_Factorization}

\end{document}